\documentclass[letterpaper]{article} 
\usepackage{aaai2026}  
\usepackage{times}  
\usepackage{helvet}  
\usepackage{courier}  
\usepackage[hyphens]{url}  
\usepackage{graphicx} 
\usepackage{natbib}  
\usepackage{caption} 
\usepackage{algorithm}
\usepackage{algorithmic}
\usepackage{amsmath}
\usepackage{amssymb}
\usepackage{multirow}
\usepackage{newfloat}
\usepackage{listings}
\DeclareCaptionStyle{ruled}{labelfont=normalfont,labelsep=colon,strut=off} 
\floatstyle{ruled}
\newfloat{listing}{tb}{lst}{}
\floatname{listing}{Listing}
\title{Beyond Binary Moral Judgment: Modeling Ethical Pluralism in AI}
\author{
    Aisha Aijaz\equalcontrib\textsuperscript{\rm 1}, Rahul Goel\equalcontrib\textsuperscript{\rm 1}, Arnav Batra\equalcontrib\textsuperscript{\rm 1}, Raghava Mutharaju\textsuperscript{\rm 2}}

\affiliations{
    \textsuperscript{\rm 1}Department of Computer Science and Engineering, IIIT Delhi, New Delhi, Delhi, 110020, India\\
    \textsuperscript{\rm 2}Mehta Family School of Data Science and AI, IIT Palakkad, Palakkad, Kerala, 678623, India
    
    aishaa@iiitd.ac.in, rahul22388@iiitd.ac.in, arnav22098@iiitd.ac.in, raghava@iitpkd.ac.in
}

\begin{document}

\maketitle

\begin{abstract}
Critical decision-making in socially consequential spaces is increasingly involving AI systems at varying capacities. Yet, despite the ubiquity of autonomous systems in domains such as medicine, law, and public interaction, most approaches to handling autonomous moral decision-making resort to scalar or binary judgments. These methods are insufficient for acceptable moral reasoning, as they provide little explanation, leaving out imperative contextual and theoretical information that must be included to support accountability. To this end, we propose a framework to model moral reasoning as a distribution over normative ethical theories or ethical pluralism, as opposed to a single ethical verdict. To achieve this, we introduce a normative ethics simplex that simplifies and integrates these theories. A benchmark of 450 cases across 15 fine-grained, sub-theories was also prepared for the purposes of stacked ensemble learning. These cases describe ethical dilemmas in natural language and have associated extracted contextual features. The implementation of the simplex was achieved via a two-stream normative-semantic architecture. This is followed by the fusion of normative information and a sequential, stacking ensemble to learn the best fit of the three broad theories: consequentialism ($\alpha$), virtue ethics ($\beta$), and deontology ($\gamma$), and the 15 subcategories. Our experiments demonstrate that the integration of contextual and normative priors with the semantic embeddings significantly improves the performance of the classification, displaying a classification accuracy of 88.89\%. To further test the model architecture, we conducted ablation studies to show that structured ethical representations contribute beyond analogical reasoning, and the chosen stacking architecture gives the best results due to the gradual learning of granularity. Ethical pluralism is also analyzed through entropy, confidence, and metric visualization. Thus, modeling ethical pluralism as a probabilistic normative distribution supports human-like moral reasoning, ethical disagreement analysis, and future alignment in AI systems.
\end{abstract}

\section{Introduction}
Artificial intelligence today participates in decisions involving healthcare, law, public policy, employment, and autonomous safety \cite{duan2019artificial}. In such settings, the usual metrics of predictive performance are not always sufficient due to the varied scale of their impact. Decision-making autonomous systems must also account for ethical parameters such as upholding rights and duties, agent intentions, consequences of actions, effects on relationships, and tradeoffs between competing ethical theories. These theories, by normative standards, include consequentialism, deontology, and virtue ethics. Yet, most approaches to the development of moral AI by design remain isolated towards any one school of thought, rather than providing a more holistic approach. In order to develop a semblance of moral reasoning in AI systems, a more structured normative attribution is required.

Human morality is contextual and grounded in duty, character, and social obligation, whereas AI operates through numerical optimization. Bridging this gap requires more than safety prompting a language model as it may provide inconsistent and unverifiable claims about what it means to be moral \cite{yahyapour2026less}. Autonomous ethical reasoning requires representation that preserves philosophical structure while remaining learnable. 

This paper approaches this challenge by moving beyond learning moral behavior from large-scale datasets and toward a more rule-based, theory-sensitive classification. The proposed model considers morality as an overlapping construct of multiple theories, emulating ethical pluralism. Instead of judging whether a case is simply morally acceptable or unacceptable, as proposed by \cite{hendrycks2020aligning}, an autonomous moral reasoner must predict which combination of normative ethical schools of thought best explains the decision and corresponding reasoning embedded in natural text. This reframing is imperative for maintaining interpretability because it distinguishes between the normative theories that may otherwise lead to similar actions, aligning with a more descriptive perspective. 

To achieve these ends, this paper proposes a hierarchical framework for ethical pluralism with increasing granularity. The moral decision of right and wrong can be linked to one or more normative ethics theories with varying degrees of membership. These theories have been subdivided into 15 subcategories in collaboration with a domain expert for a more fine-grained perspective. Next, to evaluate this framework, we constructed a benchmark dataset of 450 real-world cases where some ethical dilemmas and contextual variables are present. We introduce a two-stream ensemble learning architecture that explicitly combines moral information, semantic representation of scenarios, and structured features. The paper shows that theory classification benefits from combining philosophical taxonomy, feature engineering, and probabilistic representation. This integration aims to bridge the gap between currently lacking systems that aim to achieve similar goals; because a model that is accurate but opaque is difficult to trust, while a model that is explainable but weakly predictive may struggle to serve as a practical computational ethics benchmark \cite{uddin2025trustworthy}.

\section{Normative Ethics Taxonomy}
\label{normative ethics taxonomy}
The task to classify moral philosophy into the three perspectives is formulated primarily based on the normative categorization of ethics. This is because most subcategories of ethics theories that are cited for dilemma resolutions are linked to one of these three \cite{kagan1992structure}. Consequentialism evaluates actions primarily through outcomes \cite{sep-consequentialism}, deontology prioritizes duties and universal constraints \cite{sep-ethics-deontological}, and virtue ethics values character, wisdom, and relational morality \cite{sep-ethics-virtue}. Therefore, these are three separate categories of ethics; however, due to the large number of parameters affecting an ethical decision, a given scenario may be linked to various degrees of each of these theories. Treating them as separate but related axes is thus important given the lexical overlap in the normative principles that govern the decision.

When modeling ethics for autonomous decisions, mere binary classification becomes insufficient due to a lack of further explanation and consistency. We do anticipate the argument of confidence, as the model may learn the rate of analogical agreement with other semantically similar cases, giving some credence to the binary decision made. In contrast, the reason for its inadequacy is epistemological. A learning model, the kind that is behind most AI systems deployed in critical decision-making scenarios today, is extremely opaque and often inconsistent. They may state that an action is unethical, but will fail to recognize why or which school of thought it conforms to specifically. Furthermore, different schools of thought may justify different actions differently. Mere semantic closeness to another known case is not enough to adequately claim clarity on ethical decision-making. These reasons motivate the operationalization of ethical decisions as a distribution over normative theories. 

To visualize this, we can consider a multi-dimensional space where the major axes are represented by the normative theories, where $\alpha$ represents consequentialist influence, $\beta$ represents virtue ethical influence, and $\gamma$ represents deontological influence. Between these axes, various regions or \textit{fuzzy subsets} may exist to indicate moral rightness, wrongness, and grayness. However, by plotting the action in this space, each action becomes associated with a tuple of the three normative theories, which essentially describes its degrees of membership within these theories, and also the confidence of rightness and wrongness. Due to the nature of this mapping exercise, every point in this three-dimensional space exists in a fuzzy but interpretable position, thus adequately representing agreeing/disagreeing theories or ethical pluralism. 

\begin{figure}[t]
    \centering
    \includegraphics[width=\linewidth]{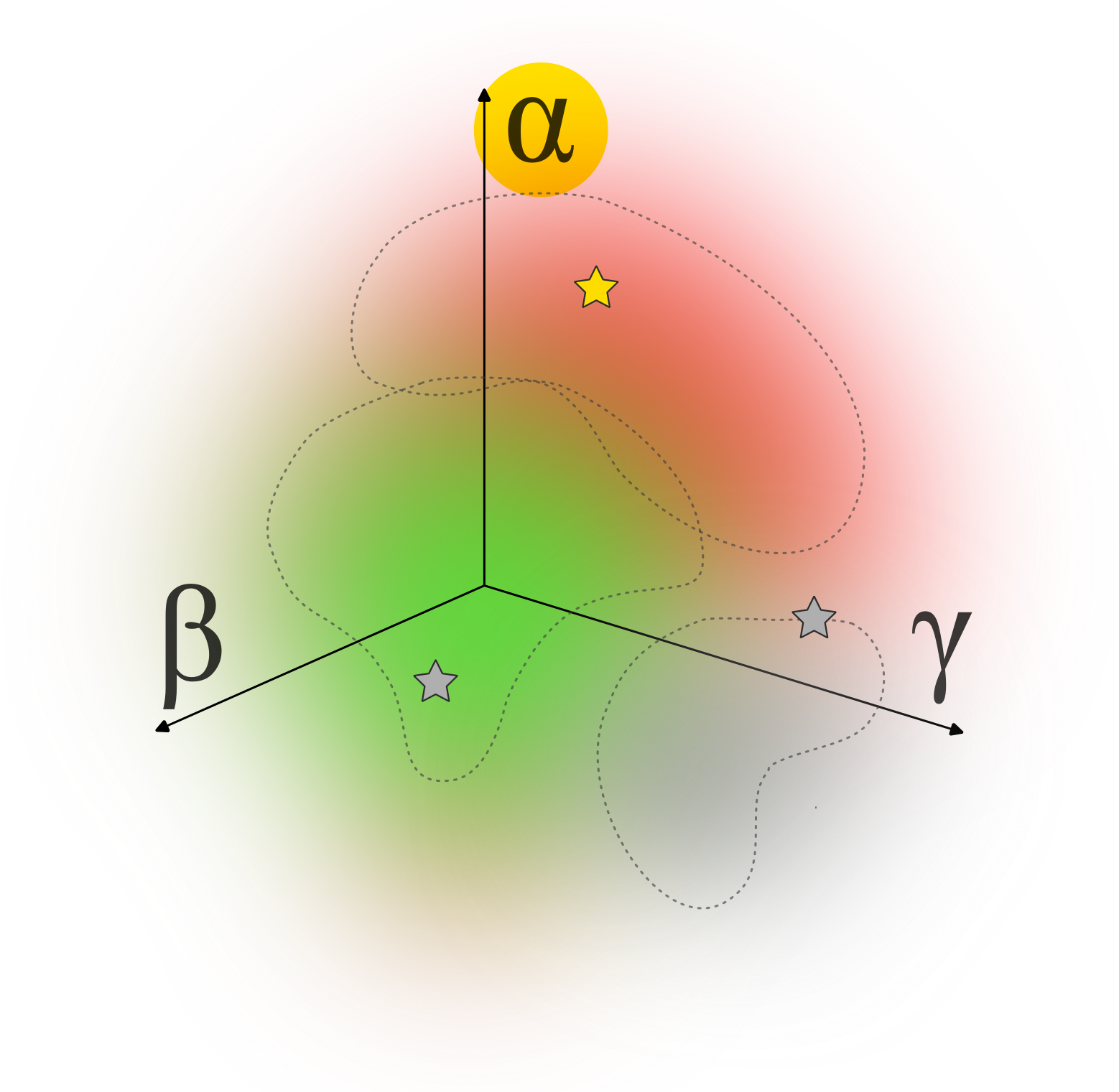}
    \caption{The diagram visualizes the fuzzy ethics subsets in a three-dimensional space, where each dimension $\alpha$, $\beta$, and $\gamma$, represent the three normative theories. This depiction visualizes competing ethical theories, or, in other words, ethical pluralism. It shows how the same action (visualized as a star) may fall into overlapping schools of thought based on the preferred ethics theory. The verdict of the action may be defined by a degree of membership of the three subsets: morally wrong (shown as the red region), morally right (green), or morally gray (gray). Here, $\alpha$ or consequentialism is chosen, under the consideration of which the action would be morally wrong. }
    \label{abcgraph}
\end{figure}

As the decision for an action with ethical ambiguity in this space must map to at least one ethical theory, we can notate the position of the action as a fuzzy partition of unity. As the action shifts from one region to the other smoothly based on varying contextual and theoretical parameters, the sum of all of its membership functions equals 1. This can be represented as a normative simplex as,

\begin{equation}
    \sum(\alpha,\beta,\gamma) = 1
\end{equation}

This is an initial representation as a coarse moral orientation space and not the final classification. We further subdivide these into ethical subtheories, as shown in table \ref{tablesubtheories}. Consequentialism is a paradigm that consists of act utilitarianism, rule utilitarianism, preference utilitarianism, negative utilitarianism, and ethical egoism \cite{sep-consequentialism}. Similarly, deontology includes Kantian ethics, prima facie duties of Ross, prima facie duties, contractualism, and rights-based deontology \cite{sep-ethics-deontological}. Virtue ethics comprises Aristotelian virtue ethics, Stoic virtue ethics, care ethics, Confucian virtue ethics, or Thomistic virtue ethics \cite{sep-ethics-virtue}. These subtheories are not exhaustive of general normative ethics theory, but were chosen to depict the pluralism of ethics and moral reasoning.

\begin{table*}[!ht]
\centering
\begin{tabular}{|p{1in}|p{1.5in}|p{4in}|}
\hline
\textbf{Normative School} & \textbf{Ethical Subtheory} & \textbf{Description} \\
\hline
\multirow{5}{*}{Consequentialism} & Act Utilitarianism & Evaluates action based on maximization of utility. \\
 & Rule Utilitarianism & Judges actions according to rules that produce the greatest good. \\
 & Preference Utilitarianism & Satisfies preferences of individuals. \\
 & Negative Utilitarianism & Prioritizes minimal suffering over maximal happiness. \\
 & Ethical Egoism & Advances self-interest of the decision-maker. \\
 \hline
\multirow{5}{*}{Deontology} & Kantian Deontology & Emphasizes universal moral duties. \\
 & Ross’s Prima Facie Duties & Proposes multiple competing duties must be balanced contextually. \\
 & Divine Command Theory & Adherence to commands that originate from divine authority. \\
 & Contractualism & Justifies actions based on mutually acceptable principles. \\
 & Rights-Based Deontology & Prioritizes protection of individual rights. \\
 \hline
\multirow{5}{*}{Virtue Ethics} & Aristotelian Virtue Ethics & Prioritizes virtuous character and practical wisdom. \\
 & Stoic Virtue Ethics & Emphasizes rational self-control and moral discipline. \\
 & Confucian Virtue Ethics & Centers morality around social harmony and respect within society. \\
 & Thomistic Virtue Ethics & Integrates Aristotelian virtues with theological principles. \\
 & Ethics of Care & Prioritizes empathy, compassion, and responsiveness to others' needs. \\
\hline
\end{tabular}
\caption{The selected subtheories for each normative ethical school of thought.}
\label{tablesubtheories}
\end{table*}

\section{Benchmark Dataset and Ethical Pluralism Representation}

A structured ethical reasoning benchmark was developed to bring across an empirical proof of concept. We handpicked 450 natural language examples from an existing large-scale dataset of ethically ambiguous natural language cases \cite{aijaz2025moral}, 30 for each subcategory of normative ethics. We designed this benchmark to include cases where ethical pluralism is also evident.

We used a taxonomical approach to define distinct features for each of these select cases, specifically, contextual variables. These are imperative, as they enhance the structure we are emphasizing when modeling ethical pluralism in these real-world cases. Morality depends on more than just the action, but also on the relational structure and context of the event in which it occurs \cite{blasi1983moral}. These contextual features provide additional inferred information such as the agents involved, both active (doer of the action) and passive (receiver of the action), the relationship between them, the moral intention, duration of the consequence, and its severity, to name a few. In table \ref{table benchmark}, the feature categories: agent, action, consequentialism, virtue ethics, and deontology contain the engineered features for this purpose.

\subsection{Benchmark Development}

\begin{figure}[!ht]
    \centering
    \includegraphics[width=\linewidth]{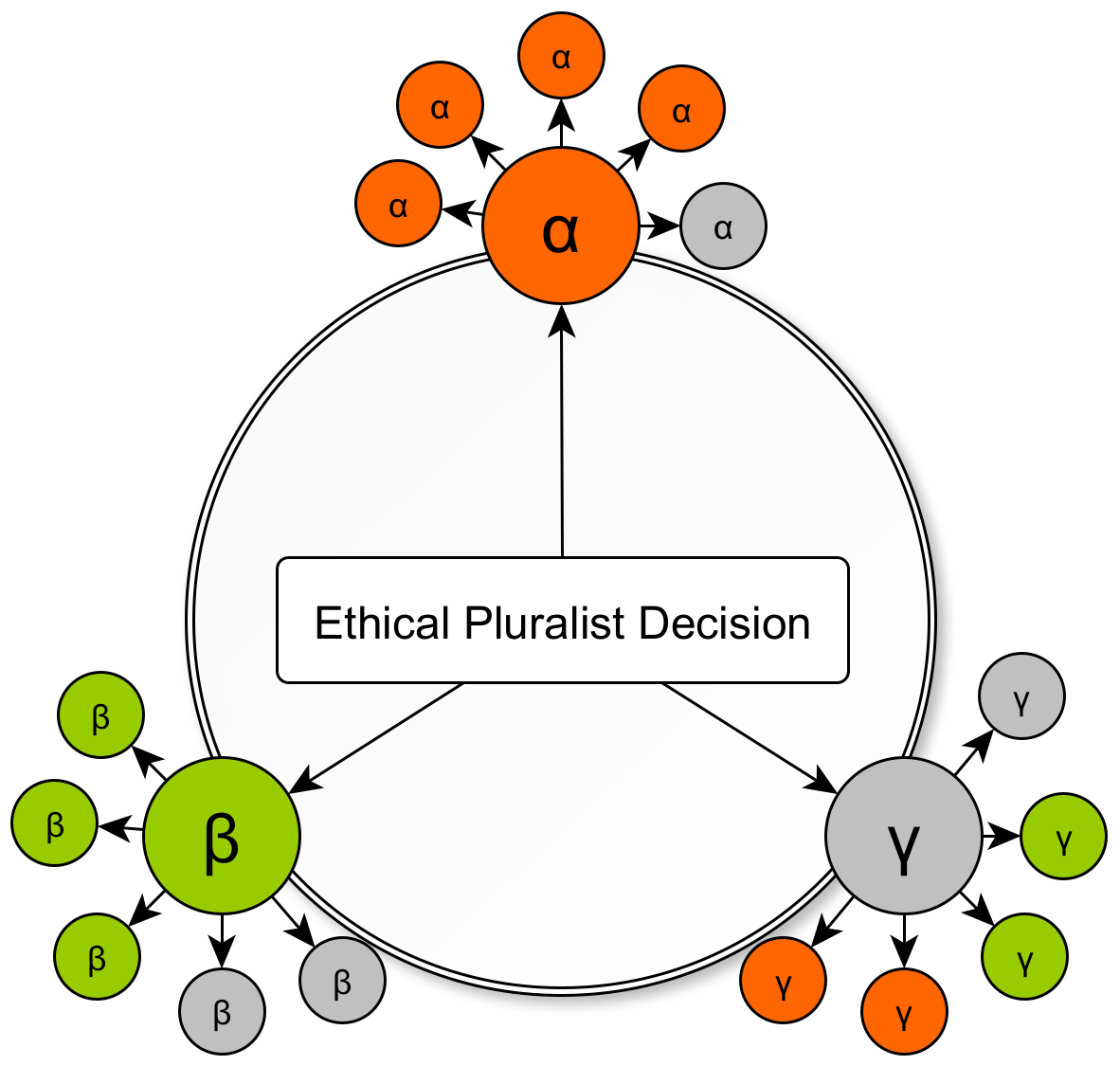}
    \caption{A visualization of an ethically plural decision. Based on varying normative theories and their subcategories, differing verdicts may be provided. Any one normative theory may be selected for the final verdict based on user preference or how the reasoner is built, however, as shown, the explanation for the decision will always be quite fine-grained with the addition of a second layer of specific theories and contextual information.}
    \label{decision}
\end{figure}

We aimed to develop a benchmark of subcategories in order to facilitate granularity when predicting the school of thought associated with a case. The granularity moves from normative ethics theory to subtheory (see figure \ref{decision}), and then to a quantification of ethical plurality. We hoped to retain real-world applicability, so we turned to a dataset that scraped first-person narrations of situations from online discussion forums to find theory-specific cases. As there is not a lot of meta-information about the raw, natural language cases with regard to the ethics involved, we inferred this with the support of a domain expert. We were also unable to find the target number of cases for all the subtheories, in which case we used Deepseek \cite{deepseekai2025deepseekv3technicalreport} to augment them based on the information we did have.

The augmentation of the cases was done via a specialized prompt, wherein the subtheory and normative theory were explained, the required template was provided, and the known examples were given. We aimed for an artificially balanced benchmark dataset to emphasize on the statistical importance of equal representation when considering a larger number of classes as opposed to simply the three normative theories. Thus, this set of 450 cases labeled with their respective subcategories acts as a structured ethical reasoning benchmark.

Ethically ambiguous cases are not only context-dependent but are also subjective, which is why structured parameters associated with them are important (see table \ref{table benchmark}). Similar actions may have different interpretations depending on the event, which is why we gathered the most recurrent and imperative information to model a general scenario with ethical ambiguity \cite{aijaz2025appleappliedethicsontology}.

\begin{table*}
\centering
\begin{tabular}{|p{1in}|p{1.5in}|p{4in}|}
\hline
\textbf{Feature \vskip 0pt Category} & \textbf{Feature} & \textbf{Description} \\
\hline
 & case\_id & Unique identifier for each case. \\
Meta-information & selftext & Raw, natural language text describing each case. \\
 & summary & Summary of each case to fit the transformer input word limit. \\
\hline
 & Active agent & Primary actor performing the action. \\
Agent & Passive agent & Individual/group affected by the action. \\
 & Agent Relationship & The kind of relationship between the interacting agents. \\
\hline
 & Action & What has been done by the active agent. \\
Action & Domain & Real-world context for applied ethics. \\
 & Ethical issue(s) & The core moral conflict raised. \\
\hline
 & Consequence & Outcome of the active agent’s action. \\
 & Severity & Magnitude of the resulting harm or benefit. \\
Consequentialism & Utility & Good, bad, or morally gray. \\
 & Duration & Temporal persistence of the outcome. \\
 \hline
Virtue Ethics & Moral intention & Underlying motive guiding the action. \\
\hline
 & Principles upheld & Moral duties, rights, or rules upheld by the action. \\
Deontology & Principles violated & Moral duties, rights, or constraints violated by the action. \\
\hline
 & Moral decision & Final moral judgment for the case. \\
Ethics Alignment  & Normative Ethics & Broad normative category.  \\
 & Ethics Subtheory & Fine-grained ethical framework. \\
\hline
\end{tabular}
\caption{The Ethics Pluralism benchmark for normative and contextual features for all 450 cases.}
\label{table benchmark}
\end{table*}

\subsection{Ethical Pluralism}
There is some discourse on a universal moral grammar \cite{mikhail2007universal}, however, in real-world situations, it is not noted for ethical ambiguity to be resolved through such a single universal moral framework. Even in general cognition, human reasoning is pluralistic, context-dependent, and shaped by many considerations; a sort of cognitive pluralism \cite{horst2024cognitive}. It would thus be naive to consider a universal, or to some extent, a binary representation of ethical cognition. However, most approaches to computational ethics rely on this assumption of scalar, one-dimensional moral thinking \cite{anderson2014geneth,awad2018moral,dehghani2008integrated}, akin to the this-or-that trolley problem \cite{thomson1984trolley}. These approaches, although valuable to the community in their own right, overlook the fact of \textit{analogical disagreement}. That is to say that similar decisions may be justified through essentially differing ethical rationales \cite{audi2006practical}. For example, withholding information may be linked to the utilitarian harm minimization, deontological upholding institutional rules, or care-oriented protection of vulnerable individuals. Thus, it would be imperative to move beyond scalar moral judgements when modeling ethical pluralism to capture multiple coexisting perspectives.

There has also been work on operationalizing moral reasoning via fuzzy interpretations of morality \cite{park2024morality,aijaz2025moral,rao2023ethicalreasoningmoralalignment}, however, we are yet to see work that leverages ethical pluralism for modeling moral ambiguity via normative ethics theories. We believe that proxy parameters such as moral principles and values do not approximate the ethical perspective with regard to an action, as much as if we adhere to established normative standards. 

This is also where we would like to emphasize on the difference between morality and ethics \cite{bauman1994morality}. Although used interchangeably in computer science discourse, most approaches focus on morality, with the consideration of domain-specific parameters and an understanding of right versus wrong. Nevertheless, ethics is a discipline, and to adhere to its philosophy, we must consider the many competing ethical schools of thought when modeling ethical ambiguity. This is why our representation space is explicitly philosophically structured. 

Thus, we decomposed the normative framework into 15 subtheories (see table \ref{tablesubtheories}). As previously mentioned, this list is not exhaustive. Any set of normative subtheories could have been selected for our experiments; however, our list is guided by coverage in philosophical discourse and computational practicality. We expect to see some overlaps, which will help us better visualize the ethical pluralism associated with these cases. The closeness in normative orientation of these theories was in several cases intentional, to include pluralistic moral reasoning. For example, rights-based deontology and kantian deontology are expected to overlap in duty-driven scenarios. Similarly, ethics of care and Aristotelian ethics may converge in some moral contexts.

This framework enables us to study ethically pluralistic reasoning as more than a problem of merely theory prediction, providing a structured representation of ethical ambiguity, overlap, and proximity. The modular streams in the learning implementation are thus able to learn distinctions between ontologically and semantically similar ethical positions, while also revealing where the overlaps in ethical subsets reduce their confidence. We see through our empirical evidence, that moral reasoning behaves not within the confines of isolated symbolic categories, but rather through occupying pluralistic positions in a shared ethics space. This finding is particularly relevant for the AI of today, as they operate in highly complex techno-social environments and within diverse communities and cultures, where competing interpretations of morality coexist. Understanding disagreement in moral positioning is thus just as imperative as a moral verdict.

\section{Methodology: Normative-Semantic Stream Architecture}
\label{Methodology}

The computational framework described for ethical pluralism has been operationalized using machine learning. We implemented various stages to make the most of the framework, namely, theory annotation using language models, normative theory, and subtheory classification using a two-stream learning architecture, and ethical pluralism analysis. Thus, we move from the philosophy of these interacting theories towards multidimensional representation, learning, interpretability, and analysis. 

The learning architecture for implementing this multidimensional normative ethics space comprises a hybrid approach that combines the normative priors with semantic language representations. These learning streams complement each other, as they present similar information to a downstream reasoner in an interpretable way. 

The normative prior stream identified the normative orientation of the cases through probabilistic ethical priors and meta-analysis of overlap and ambiguity. The scores of $\alpha$, $\beta$, and $\gamma$, provide an inferred normative representation of the ethicality of each case from the benchmark dataset. We also calculated the score margins, entropies, and pairwise ratios of the top two normative alignments.

Following this, we focused on the representation of the raw text and structured context of the cases. This involved converting each natural language case to a high-dimensional vector using Triple-BERT embeddings \cite{talaat2025novel}. The concatenated embeddings from three transformer models presented a supervector of 1920 dimensions. This supervector, in conjunction with one-hot encoded (categorized) contextual features, is the output of the semantic contextual stream. 

This set of representations of ethically ambiguous cases are subsequently combined and processed via a stacked ensemble classifier for more fine-grained ethical-theory classification. Our objective through the use of this multi-level architecture is not merely to classify moral scenarios, but also to learn how ethical theories overlap, diverge, and interact within a shared representation space. This section describes our methodology in more detail.

\subsection{Normative Prior Stream}
In order to computationally model ethical pluralism, the first task was to introduce a modeling framework for the normative ethics theories. This involved positioning each case with a probabilistic alignment over the three theories: consequentialism, virtue ethics, and deontology. This framework, due to its non-rigid structuring, allows the cases from the developed benchmark to be placed in proximity to multiple ethical perspectives simultaneously. The motivation for this was to emulate real-world moral reasoning, which is often ambiguous, context-sensitive, and supported by pluralistic ethical notions. 

\paragraph{Scores of Theory Alignment.} As mentioned in section \ref{normative ethics taxonomy}, the relative influence of the three normative schools is represented as a partition of unity, where the sum of \textit{scores of alignment} of each case in the benchmark equals 1. 

\begin{equation}
    <\alpha,\beta,\gamma> \in \mathbb{N}
\end{equation}

We used the open-source DeepSeek V3 \cite{deepseekai2025deepseekv3technicalreport} to evaluate these scores for each case. Upon manual expert analysis of the scores received, this LLM had the closest rate of agreement to human annotators. Based on these findings and for our experimentation, we consider the large-scale annotation of the LLM for the benchmark as the true values for our classification task. These values may be considered as \textit{ethical priors} that describe the moral positioning of the case in the 3-dimensional and continuous ethics space. These scores, however, are not the final classification of ethics theory; instead acting as a broad characterization for the subsequent steps in the methodology. Therefore, cases that exhibited higher values of $\alpha$ were strongly aligned with consequentialism, whereas higher $\beta$ values meant character or care-oriented case, and high $\gamma$ indicated duty or principle oriented ethical reasoning.

\paragraph{Scores of Ethical Plurality.} The second level of normative theory modeling involved five subtheories for each category. Apart from the $\alpha$, $\beta$, and $\gamma$ scores and the contextual features outlined in table \ref{table benchmark}, we also engineered features to capture \textit{normative theory ambiguity and overlap} between these subtheories. We identified the dominant ethical schools, pairwise score ratios, score margins, and entropy-based measures of uncertainty. We learned from the calculation of normalized entropy between the benchmark cases that those with concentrated distributions exhibited stronger alignment with one normative school of thought, thus displaying greater confidence, whereas those with more spread out distributions indicated overlapping and lower confidence. 

\begin{equation}
    <M, En, T_1/T_2, T_1> \in \mathbb{P}
\end{equation}

This conceptual formulation of a 3D space with the three normative theories as axes allows the cases to occupy positions in the fuzzy subsets with varying degrees of membership. The ethical priors and this probabilistic representation of normative ethics form the first stream of our two-stream normative-semantic architecture.

\subsection{Semantic-Contextual Stream}

Ethical reasoning with increased granularity depends on semantic nuance and contextual information. This stream provides the information needed to process these representations in two steps. 

\paragraph{Semantic Embeddings.} Representing semantic information for the cases involved a further two steps. As previously mentioned, we leveraged natural language processing to generate a 1920D supervector that directly represented the raw cases via vector embeddings. This was done using the triple BERT transformer, which contains three separate sentence transformers, and optimizes the triplet loss function \cite{dong2018triplet}. We chose the following transformers for our embedding task:
\begin{enumerate}
    \item \texttt{all-MiniLM-L6-v2}, maps sentences to a 384-dimensional dense vector space. Trained on 1B sentence pairs. We used this as it creates a shorter, more efficient embedding but uses a longer input. Therefore, it gives us an initial mapping of the general content of each case.
    \item \texttt{all-distilRoBERTa-v1}, maps sentences to a 768-dimensional dense vector space. Trained on 1B sentence pairs. This gives us a longer embedding from a shorter input, thus working with the summarized cases to be less than the 128 word limit. This provides us a more detailed representation of the most pertinent information.
    \item \texttt{multi-qa-mpnet-base-dot-v1}, also maps sentences to a 768-dimensional dense vector space. It is trained on 215M question-answer pairs. This transformer adds an additional layer of back-and-forth reasoning, which gives more structured information to position each case in the 3-dimensional normative ethics region. 
\end{enumerate}

Therefore, with the concatenated vectors of the three sentence transformers, we get a 1920-dimensional supervector to represent each case. This technique emphasizes semantic representation diversity across the three transformers, thus reducing representational bias and enhancing robust embeddings across varying cases.

\begin{equation}
    <E_1,E_2,E_3> \in \mathbb{SV}
\end{equation}

\paragraph{Contextual Categories.} A second set of parameters needed for the contextual representation were the one-hot encoded values for features with predefined classes. These include the characteristics of consequence: severity, duration, and utility, moral intention, and principles upheld and violated. 

\begin{equation}
    <C_{Sev},C_{Dur},C_{Ut},MI,P_{up},P_{vi}> \in \mathbb{C}
\end{equation}

Although the triple BERT embeddings provide a highly nuanced representation of the benchmark cases in order to find the slight differences between them, these contextual parameters augment them by providing explicit contextual grounding beyond linguistic similarity. Overall, this stream captures the case context linguistically, whereas the normative prior stream captures alignment with specific ethical theories.

\subsection{Stacked Learner and Meta-learner}

Next, the ethics priors and semantic-contextual representations are combined to provide a unified feature space that would enable a learning model to reason over normative alignment and narrative semantics. For this, the learning model implemented is a stacked ensemble of three base learners \cite{dey2023ensemble}, which, upon experimentation of choice of models, gives us the best results when using the bagging-boosting-stacking method as is used for other nuanced classification tasks \cite{ribeiro2020ensemble,wen2020coastal}. 

\paragraph{Bagging.} Random Forest is a commonly used bagging technique that creates random subsets of the dataset, trains a classifier on each of those subsets, and aggregates the results. This technique mimics a kind of sampling, representing the many subtheories in the benchmark and providing results for each run that are independent of each other.  

\paragraph{Boosting.} XGBoost is a powerful gradient boosting algorithm that is very good at performing multi-level classification with many labels. Since we perform a 15-way classification exercise, the use of XGBoost was fitting in abstract settings.

\paragraph{Stacking.} We used a linear Support Vector Machine, which was used to provide linear separation in fuzzy, high-dimensional spaces. This was a fitting choice, as for a high-dimensional space (over 1920 dimensions), a dataset with only 450 instances makes it very simple for a linear classifier to draw hyperplanes separating the 15 subtheories. This model is also immune to overfitting due to the regularization of weights. Furthermore, this is a highly time and space-efficient algorithm. 

\paragraph{Meta-Learner}
Instead of implementing a discrete vote to decide on the classification of theories from the base learner, we used a second XGBoost to combine the predictions of the three base learners for the optimal final classification. We found that providing a probabilistic reassurance aided our efforts towards interpretability, which is something we prioritized for ethical pluralism modeling. 

This architecture allowed us to aggregate various decision patterns to improve robustness and performance. Furthermore, the computational complexity of these models is significantly lower than that of more demanding specialized ML algorithms. Although it may seem like a complex architecture, the simple concatenation of otherwise individually \textit{weak} learning models together provides a highly sensitive learner, the kind that is apt for a linguistically sensitive classification task such as this. 

\section{Results}
In this section, we outline the experiments done to achieve two major research objectives, namely, (a) to predict fine-grained normative ethical theories, and (b) to quantify and analyze ethical pluralism. 

\subsection{Theory Classification}
Our proposed Normative-Semantic Stream Architecture achieves an exact-match accuracy of 88.89\% and a macro F1 score of 88.78\% when predicting one of the 15 subtheories. As with our temperature scaling, our choice of metrics is also conservative and leaves little room for the model to be creative. This is due to the nature of the task, where confidence is low, and the performance is evaluated based on certainty. We also calculated the macro F1 scores to ensure that the classification mirrors our initial setting of a balanced dataset by treating all classes equally.

\subsection{Ablation Study}

Since we are using various combinations at different steps, it was imperative to conduct an ablation study to see how the different elements in our normative-semantic stream architecture contribute to the classification task (see Table \ref{tab:ablation_results}). The proposed Normative-Semantic Stream architecture achieved the strongest classification performance of 89\%, which dropped to 85\% upon removing the normative priors, further down to 81\% when removing context, and merely 77\% with only semantic representation via embeddings. This indicates that the probabilistic ethics priors as well as the contextual parameters contribute meaningful structure and reasoning ability beyond solely linguistic representation. This bolsters the evidence of ethical plurality and the pertinence of contextual and normative priors for more accurate classification.

We did a similar study to test the need for a 1920D supervector. Upon removing the individual sentence transformers, we noticed a measurable degradation of performance, which showed that our initial assumption of capturing layers in semantic reasoning via diverse transformers was well-founded (see Table \ref{transformer ablation}). 

From this study we concluded that the best performance for subtheory classification was observed when there was active interaction between semantic nuance, rigid ethics structure, and dynamic context with diverse layers in linguistic representation and learning aspects. 

\begin{table}[!ht]
    \centering
    \begin{tabular}{|p{1.4in}|c|c|}
        \hline
        \textbf{Model} & \textbf{EM Accuracy} & \textbf{Macro F1} \\
        \hline
        \textbf{Normative-Semantic} ($\mathbb{N+P,SV+C}$) & \textbf{0.8889} & \textbf{0.8878} \\
        Only $\mathbb{SV+C}$ & 0.8556 & 0.8579 \\
        Only $\mathbb{N+P,SV}$ & 0.8111 & 0.8075 \\
        Only Embeddings ($\mathbb{SV}$) & 0.7778 & 0.7700 \\
        \hline
    \end{tabular}
    \caption{Feature ablation results.}
    \label{tab:ablation_results}
\end{table}

\begin{table}
\centering
\begin{tabular}{|p{1.4in}|c|c|} 
\hline
\textbf{Transformer \vskip 0pt Configuration} & \textbf{EM Accuracy} & \textbf{Macro F1} \\
\hline
\textbf{TripleBERT} & \textbf{0.8889} & \textbf{0.8878} \\
Without MiniLM & 0.8667 & 0.8689 \\
Without RoBERTa & 0.8778 & 0.878 \\
Without MPNet & 0.8778 & 0.8745 \\
\hline
\end{tabular}
\caption{Transformer Model Ablation Results.}
\label{transformer ablation}
\end{table}

\subsection{Ethical Pluralism Analysis}

The objective of this paper was to model ethical pluralism through a multi-dimensional representation task. We performed fine-grained classification to see how well a learning model could differentiate between very similar normative ethics subtheories. Upon much experimentation and representation exercises for natural language cases with ethical ambiguity, we arrived at the best performing ensemble architecture, which gave an EM accuracy of 89\%. This low accuracy can be attributed to the lower confidence of the model, and is what makes this problem more interesting.

Ethical pluralism is a continuous, overlapping phenomenon, and we expect machine learning models to find trouble in providing optimal or highly confident results. In this subsection, we analyze and visualize this behavior through an extensive ethical overlap analysis. From the probability distributions for the cases across the 15 subtheories, we computed entropy, confidence margins, and theory-overlap scores. 

\begin{figure}
    \centering
    \includegraphics[width=\linewidth]{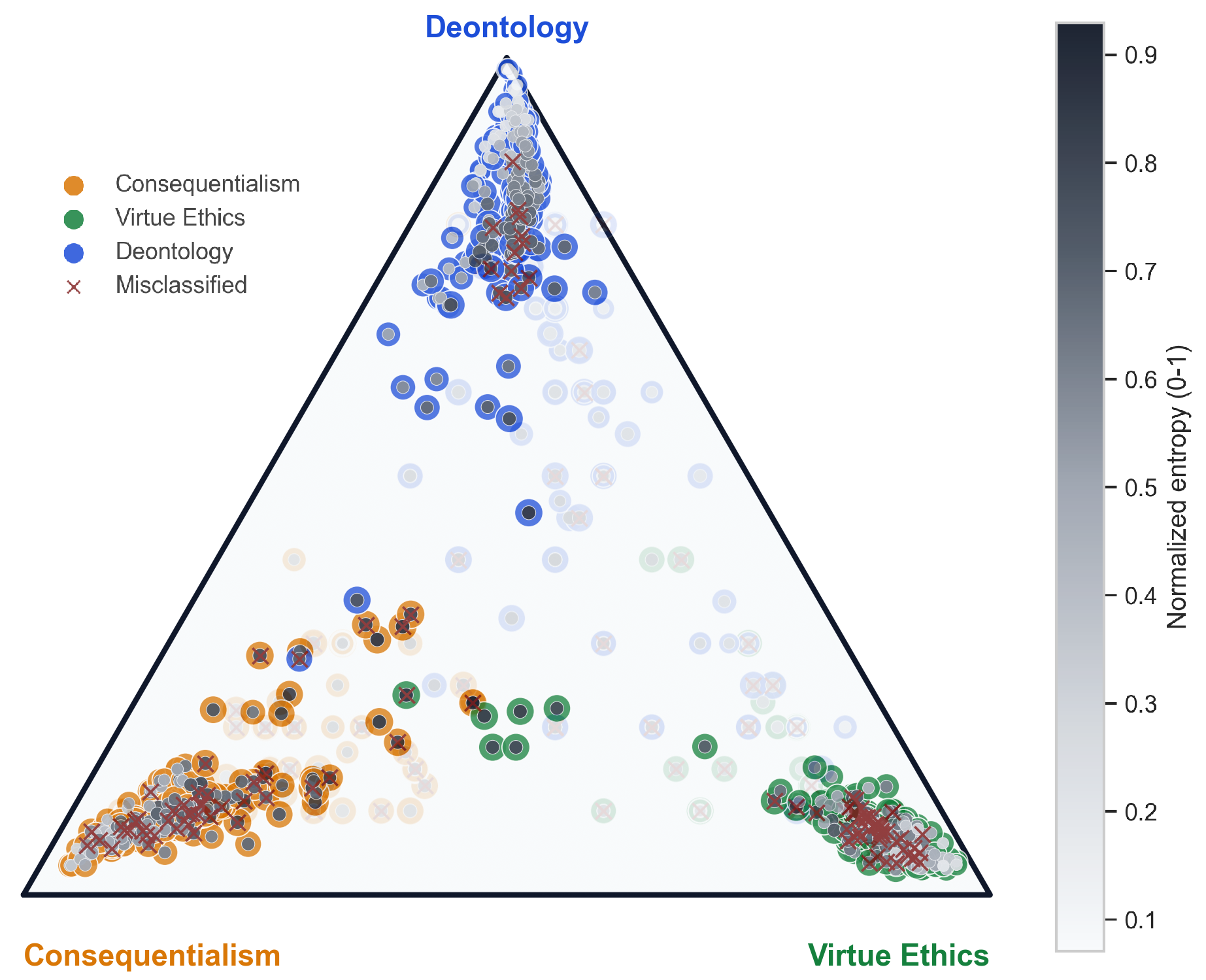}
    \caption{The Normative Simplex across the three normative ethics theories. The fill of each case shows decreasing entropy towards the corners of the simplex and higher entropy towards the center. This visualization shows the entropy of the model classifications after normalization superimposed over the visualization before normalization. Thus the difference in entropies may be seen.}
    \label{superimp}
\end{figure}

\paragraph{Entropy.} This metric is used as a measure of normative ambiguity. We noted a higher entropy in the model upon observing the results of the stacked ensemble, which required temperature scaling normalization to reduce the entropy. Using $T=0.6$, a cooler temperature for normalization, we were able to reduce the entropy by a third, from 1.66 to 0.61. We visualize the normative simplex that represents the partition of unity in the fuzzy ethics subset in Figure \ref{superimp}.



\begin{figure}[ht]
    \centering
    \includegraphics[width=\linewidth]{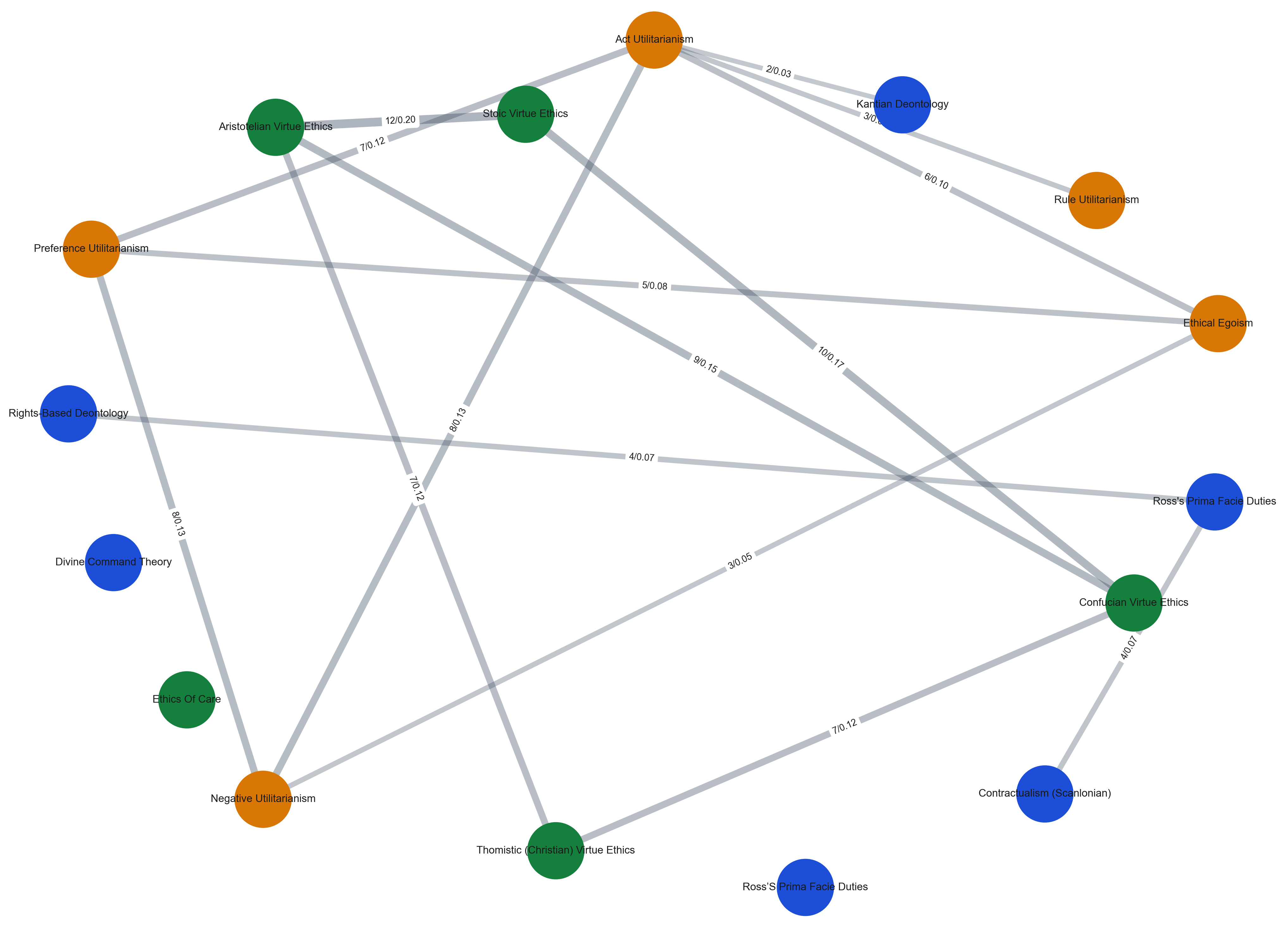}
    \caption{Bridge Theories that show pairwise confusion frequencies. The model found most difficultly in classifying cases from virtue ethics subtheories (Aristotlean, Confucian, and Stoic Virtue Ethics), and some Consequentialist subtheories (Act Utilitarianism and Preference Utilitarianism). The most distinct subtheories with no confusion were Divine Command and Prima Facie Duties. An interesting insight from this graph is that most bridge theories lie within the same normative categories.}
    \label{theory overlap}
\end{figure}

\paragraph{Overlap Scores.} In addition to looking at a general chaos representation through entropy, we also want to quantify the relationships between the 15 ethics subtheories. For this, we conducted a bidirectional overlap study across all subtheories using pairwise confusion frequencies. We observed several bridge theories that were learned by our architecture, which may or may not be otherwise very distinct. Those theories with high confusion frequencies showed up more in low confidence, ambiguous classifications. See
Figure \ref{theory overlap}.

\begin{figure}[ht]
    \centering
    \includegraphics[width=\linewidth]{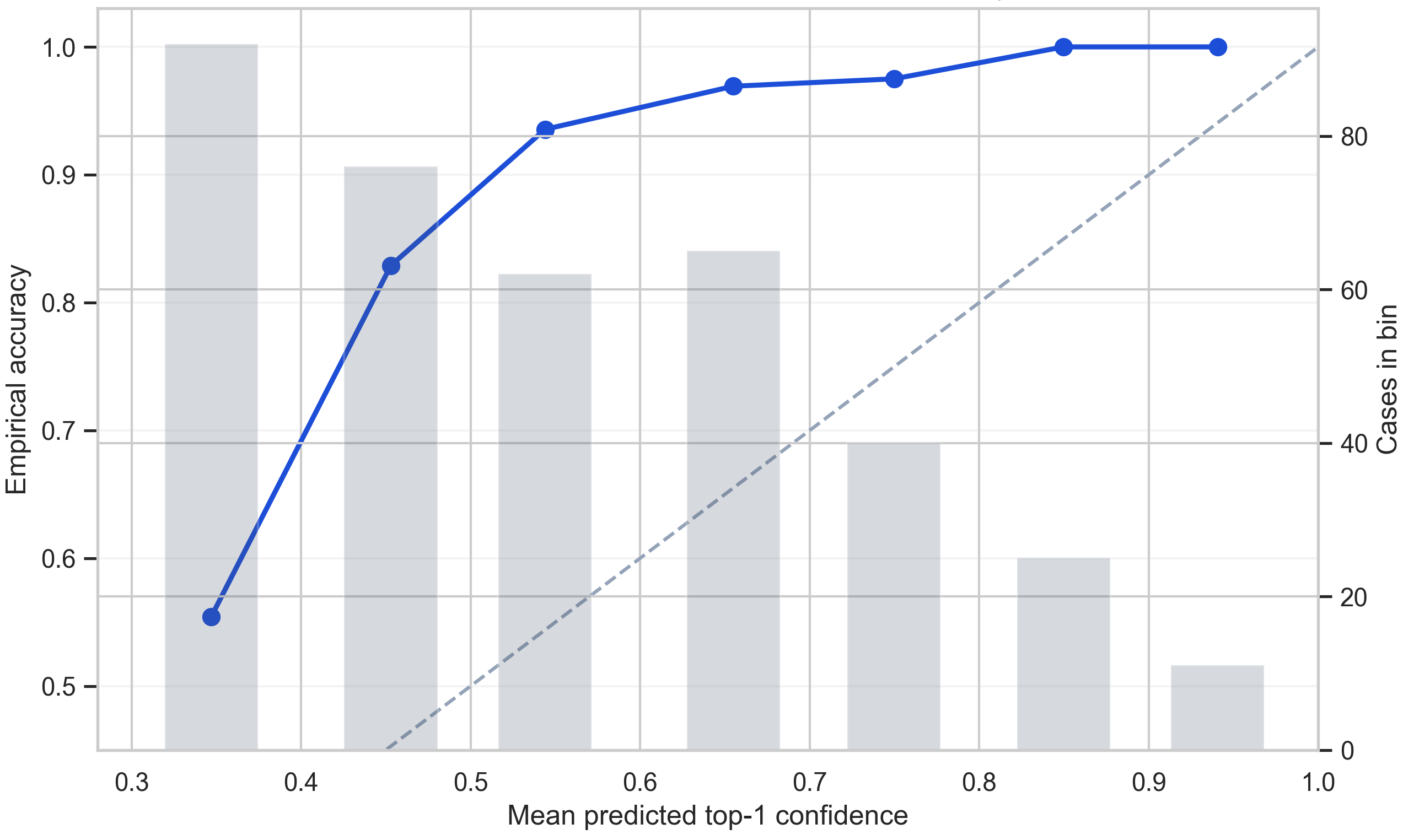}
    \caption{Confidence Stratification Curve.}
    \label{Confidence}
\end{figure}

\paragraph{Confidence.} The reliability of the model was evaluated using a confidence-stratification analysis (See Figure \ref{Confidence}). We can see that the model is able to give an above 70\% top-1 confidence for only about 15\% of the cases. It is noted that with increasing confidence, we get increasing empirical accuracy. On the other hand, low-confidence indicates greater ethical ambiguity or overlapping cases. 

This predictive uncertainty is also imperative in understanding the representation of ethical plurality in the feature space. Instead of simply classifying the cases into deterministic moral verdicts in highly ambiguous scenarios, we can reveal the uncertainty through these measures, as well as previously discussed probability distributions and entropy. A threshold of uncertainty may thus be implemented to recognize cases that may require further human deliberation and contextual oversight.

\paragraph{Supervector Visualization.} To help us place the benchmark cases in the actual feature space, as we did for representation purposes in Figure \ref{abcgraph}, we had to visualize the 1920D supervectors for each of the cases. Although this is impossible to do in 2D, we were able to project the supervectors into lower-dimensions using Uniform Manifold Approximation and Projection (UMAP) \cite{mcinnes2018umap} and simplex-based geometric visualization. This is the probabilistic ethical coordinate space defined by the normative ethics theories, mapped to a 2-dimensional UMAP. The cases associated with the same normative theories tend to cluster together, showing that the architecture inclines towards the normative ethical structure. Therefore, ethical reasoning, due to its pluralistic nature, benefits from being represented in a probabilistic or fuzzy region, rather than a three-way classification with more discrete labels. The learned geometry shows observable evidence towards these fuzzy regions and the overlapping normative structures within ethical reasoning.

\begin{figure}[ht]
    \centering
    \includegraphics[width=\linewidth]{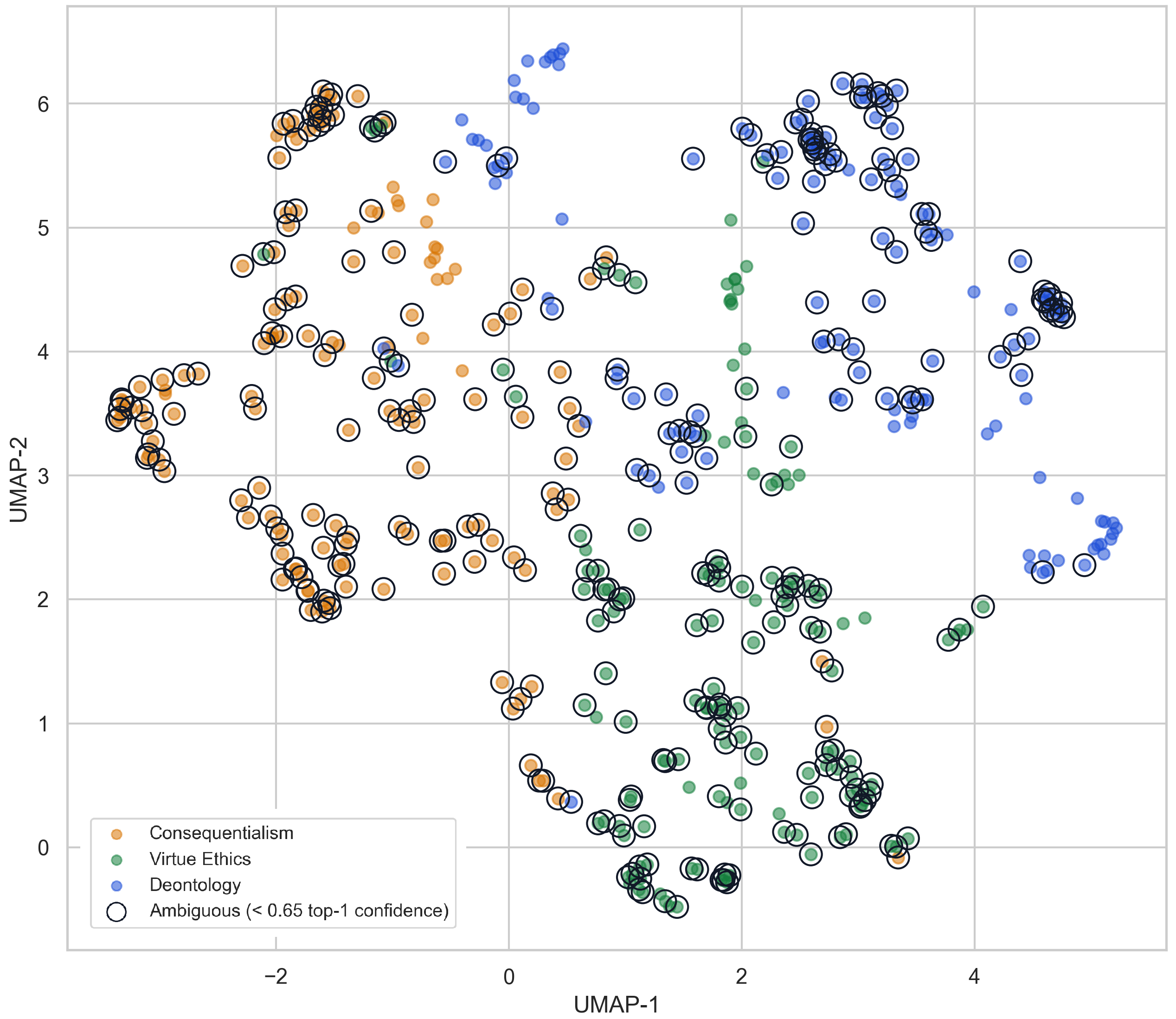}
    \caption{2D UMAP for Supervector visualization. Each point is a continuous representation of the case which includes ethical nuance and contextual parameters. The circled cases are marked as ambiguous by the classifier architecture.}
    \label{umap}
\end{figure}

\section{Related Work}
Although this area of research is highly imperative, there is less proportional discourse on developing AI systems that are able to reason ethically over their decisions. Nevertheless, there are works that provide valuable contributions to the modeling of ethics theories for human-aligned reasoning. In the work by Park et. al. \cite{park2024morality}, ethical pluralism is also modeled as embeddings, albeit using a contrastive learning approach to show that such ethical cases can be represented in this way. Although this paper sets up a valuable path for future ethical modeling research, it misses out on a normative ethical structure and is instead grounded in latent semantic structure only. Our work offers a hybrid philosophical and semantic approach, which also takes into account various subtheories for fine-grained reasoning, as well as dynamic contextual aspects for additional nuance. We also present an ethical overlap analysis via bridge theories, confidence, and entropy to also quantify ethical pluralism. Anderson et. al. \cite{anderson2008ethical}, on the other hand, represent ethics merely through symbolic and rule-based architectures, while our experiments show the importance of a more pluralistic representation. Hendrycks et. al. \cite{hendrycks2020aligning} present an ETHICS dataset for human alignment, but vastly differs from our method, as they do not consider normative classifications at all, instead labeling their cases as \textit{morally acceptable} or \textit{unacceptable}. Various other approaches towards integrating ethical reasoning into AI systems exist in the literature \cite{rao2023ethicalreasoningmoralalignment,bai2022constitutionalaiharmlessnessai,awad2018moral,dehghani2008integrated}; however, we would argue that the task primarily necessitates the development of pluralistic ethics representation, followed by other methods such as prompting, crowd-sourcing, analogical reasoning, and post-hoc decision flipping.

\section{Conclusion}
Our paper models ethical pluralism for representation in AI through a 2-stream architecture that combines normative ethics and semantic context. This was implemented via an increasingly granular approach to ethics theory, which represented the normative schools, consequentialism, virtue ethics, and deontology, as a simplex partition of unity. This indicates that the model will collapse to at least one decision in the three-way classifier problem. To test the architecture, we chose a subset of an established Moral Decision Dataset to manually curate a benchmark dataset of 15 normative subtheories, with 5 associated with each broader school. We chose a stacking ensemble for the learning model to emulate varied reasoning pathways, and aggregated the results of the chosen base learners using a powerful metalearner for optimal theory classification. Our results show that such structured normative information and dynamic contextual parameters are necessary to the representation exercise, as the model behaves pluralistically when left with merely semantic embeddings. This ethical pluralism has also been quantified by various metrics, such as entropy, confidence, and UMAP visualization. 

Although our cases are taken from real-world scenarios individually, we would like to remind readers that our benchmark dataset is relatively small and balanced, and does not reflect real-world environments. This was an active choice to merely model ethical pluralism and test best performing architecture configurations in a limited setting. Secondly, the annotation of the scores of alignment relies on LLM-generated priors. As mentioned, these have been verified by agreement rates with human annotators and were generated via expert-guided prompts, but still remain at risk of bias and deviation from a perhaps wider settlement. Thirdly, the selected 15 subtheories are not an exhaustive list and do not represent all demographic inclinations of normative ethics. These were chosen to support the benchmark creation and reflect mainstream philosophical discourse. Lastly, the framework does not perform adaptive conflict resolution, only collapsing to the best-fit probability for classification. 

Future work may extend the benchmark to include additional diverse cases and theories inclined towards various communities and cultures. The quantification of plurality in these specific regions may provide interesting insights and divergence from conventional normative ethics. Other tasks that may benefit from this representative modeling, apart from classification, include dilemma resolution, disagreement detection, uncertainty estimation, and human deliberation thresholding. This work thus shifts the perspective from verdict-based reasoning towards structuring ethical plurality, thus depicting moral reasoning more inclined to the real-world.

\bibliography{aaai2026}

\end{document}